# Rule-Guided Joint Embedding Learning over Knowledge Graphs

Qisong Li[1], Ji Lin[1], Sijia Wei[2], Neng Liu[3]

[1]School of Computer Science, Wuhan University of Technology
[2]School of Computer Science, Jianghan University
[3]School of Computer Science, Wuhan University of Science and Technology

**Abstract**

Recent studies on knowledge graph embedding focus on mapping entities and relations into low-dimensional vector spaces. While most existing models primarily exploit structural information, knowledge graphs also contain rich contextual and textual information that can enhance embedding effectiveness. In this work, we propose a novel model that integrates both contextual and textual signals into entity and relation embeddings through a graph convolutional network. To better utilize context, we introduce two metrics: confidence, computed via a rule-based method, and relatedness, derived from textual representations. These metrics enable more precise weighting of contextual information during embedding learning. Extensive experiments on two widely used benchmark datasets demonstrate the effectiveness of our approach, showing consistent improvements over strong baselines.

## 1 Introduction

In recent years, knowledge graphs have been widely used to organize and publish structured data in various domains due to their advantages of high expressive power, low ambiguity, uniformity of schema, and support for reasoning. Typically, a knowledge graph consists of entities, their attributes, and relationships between entities. For example, it may contain the entity China, the relationship capital, and the entity attribute "China". The base composition of a knowledge graph is a triple that describes the relationship between two entities or the relationship between an entity and its attributes, e.g. (China, Capital, Beijing), (China, English label, "China").

At present, knowledge map has been widely used in tasks such as intelligent question answering [1], recommendation system [2] and information retrieval [3], and its outstanding performance has been widely concerned by both academia and industry [4]. However, while benefiting from the rich information contained in knowledge map, its huge scale and data sparseness have also brought challenges to the application of knowledge map. For example, open domain knowledge maps such as Freebase [5], Yago [6] and dbpedia [7] usually contain millions of entities and hundreds of millions of triples describing the relationship between entities. Traditional graph algorithms like subgraph matching often struggle with computational efficiency when applied to large-scale knowledge maps. To address this, researchers have developed a knowledge graph embedding learning model. This model transforms the knowledge graph into a continuous, low-dimensional vector space, enabling the efficient learning of embedding representations for entities and relationships.

By designing a specific representation learning mechanism, information such as the structure and semantics of the knowledge map can be encoded in the learned embedded representation. On the one hand, large-scale knowledge maps originally needed to be frequently visited, such as structured query. Construction) [9], Logical Query Pro-Cessing [10] and query relaxation) [11] can all be completed by numerical calculation in the learned embedded representation space, which greatly improves the efficiency. On the other hand, the embedding learning of knowledge map provides a method to extract and efficiently represent the feature information of knowledge map, which is similar to word embedding, which is widely used in the field of natural language processing, and the embedding representation of knowledge map also provides great convenience for deep learning based on knowledge map.

Most of the existing knowledge map embedding learning models only pay attention to the structural information represented by triplets in the knowledge map. For example, Bordes et al. put forward the TransE model based on translation mechanism [12], whose target tasks are link prediction and link prediction.Triple classification, in a nutshell, is to judge whether there is a certain relationship between two given entities in the knowledge map. Therefore, the TransE model only pays attention to the encoding of a single triple structure information by the learned embedding representation, which simplifies the knowledge map into a finite set of unrelated triples in the embedding learning process. Therefore, Transition and its subsequent improved models [13-16] have very weak coding ability for contextual information in knowledge maps, so it is difficult to be applied to semantically related tasks. To solve this problem, some embedded representation models based on contextual information have been proposed one after another. However, they still only pay attention to contextual information represented by subgraphs, paths and other structures in knowledge maps. For example, When learning the embedded representation of the entity Beijing in Figure 1, the above method only pays attention to the triplets (China, capital, Beijing) and (Beijing, located in North China) which describe the relationship between entities, and ignores the text information such as the introduction of Beijing and English labels. Obviously, the lack of text information limits the expression of semantic information in the learned embedded representation.

In order to solve this problem, this paper proposes a rule-guided knowledge map joint embedding learning model. Inspired by the graph convolution network, the model firstly encodes the context information of the entity in the knowledge graph into the embedded representation of the entity through multi-relational graph convolution. Different from the work of Vashishth and others, this paper holds that multiple pieces of context information of an entity should have different degrees of importance, and the degree of importance of a certain piece of context information depends on two factors: the confidence of the piece of context information and its relevance to the entity. Therefore, this paper puts forward a simple and effective rule to guide the calculation of the confidence of context information, and based on the text information representation in the knowledge map, puts forward a calculation method of the correlation degree between entities and their context information. Finally, the model integrates the embedded representation encoded by graph convolution network with the vector representation of text information, and takes the result of link prediction task as the training goal to learn the embedded representation of entities and relationships in knowledge map.

The contribution of this paper is mainly reflected in three aspects:

1) Based on the graph convolution network, an embedding representation learning model guided by rules is innovatively proposed, which considers the context information and text information in the knowledge map jointly.

2) According to the importance of context information in graph volume product, a new method is proposed to calculate the confidence and correlation of single context information by applying rules and text information in knowledge map.

3) Experiments are carried out on benchmark data sets and compared with related knowledge map embedding learning methods. The experimental results verify the effectiveness of this model.

## 2 Related work

In this section, the knowledge map embedding learning model related to this work is introduced. Because the model proposed in this paper is based on graph neural network, the knowledge map embedding learning model based on graph neural network and other non-graph neural networks are introduced respectively.

### 2.1 Model based on graph neural network

The models based on graph neural network mainly include R-GCN [22] W-GCN [23], COMPGCN [21] and so on. This kind of model usually uses the graph convolution network as the encoder to encode the graph structure data, and combines with the corresponding decoder to perform tasks such as link prediction and node classification on the knowledge graph. In R-GCN, the characteristics of nodes and relationships in each layer of the network are calculated by using the weight matrix, and transmitted to the subsequent network layers through domain aggregation. Specifically, R-GCN uses base decomposition and block diagonal decomposition to construct the weight matrix of a specific relationship, so as to deal with different types of neighbor relationships, fuse them with neighbor node information, and transmit them to the target entity for updating. W-GCN assigns learnable weight parameters to each weight matrix in the process of graph-volume-product network aggregation, so that the model can obtain a better entity embedding representation. CompGCN proposes a domain information aggregation method for the central node, using a variety of "entities" in theory.

### 2.2 Non-graph neural network model

There are various types of embedding models for non-graph neural networks. These include translation-based models, such as TransE [12] and its subsequent extensions TransH [13], TransR [14], and TransD [15]; rule-based models, such as Neural-LP [16] and TILP series [17–19]; context-based models, such as GAKE [20] and RDF2Vec [21]; and tensor decomposition–based models, such as ComplEx [24] and RESCAL [25].

The translation mechanism of TransE is relatively simple, so it can be efficiently applied to large-scale knowledge maps, but at the same time, it limits the expressive ability of its models, making it difficult to deal with complex relationships of one-to-many, many-to-one and many-to-many types [14]. In order to solve this problem, some models with more complicated translation mechanisms have been proposed after TransE. For example, TransH [13] designs the translation mechanism relative to the hyperplane space of the relations in the given triplet, while TransR [14] learns an extra matrix for each relation in the knowledge map, with which the head and tail entities are mapped into the corresponding relation vector space through linear transformation, and then calculates the loss value of its translation mechanism.

## 3 Joint Embedding Representation Learning

In this section, we first provide a formal definition of the knowledgegraphembedding learning problem, introduce the notationofrelated concepts, and then introduce the proposed rule-guided joint embedding learning model indetail.

### 3.1 Problem Definition

In this paper, the knowledge map is expressed as $\mathcal{G} = (\mathcal{E}, \mathcal{R})$, where $\mathcal{E}, \mathcal{R}$ respectively represent the entity and relation set in the knowledge map. For a triple $(e_{\mathrm{h}}, r, e_{\mathrm{t}}) \in \mathcal{G}$, in which the head and tail entities all belong to the entity set, that is, $e_{\mathrm{h}}, e_{\mathrm{t}} \in \mathcal{E}$, and the relationship belongs to the relationship set, that is, $r \in \mathcal{R}$. The embedding learning problem of knowledge map is to learn the vector representation e of any entity $e \in \mathcal{E}$ and any relationship $r \in \mathcal{R}$ in a given knowledge map $\mathcal{G}$, $r \in \mathbb{R}^d$, where is the dimension represented by $d$ embedding. A test task evaluates the learned embedded representation, this task may include two scenarios: given entity $e \in \mathcal{E}$ and relation $r \in \mathcal{R}$, based on them embedding represents $\boldsymbol{e}, \boldsymbol{r} \in \mathbb{R}^d$, and predicts another entity $e' \in \mathcal{E}$, so that there are three groups $(e, r, e') \in \mathcal{G}$ or $(e', r, e) \in \mathcal{G}$; Or given two entities $e, e' \in \mathcal{E}$, based on their embedded representations $\boldsymbol{e}, \boldsymbol{e}' \in \mathbb{R}^d$, predict a relation $r \in \mathcal{R}$, so that triple $(e, r, e') \in \mathcal{G}$ exists or $(e', r, e) \in \mathcal{G}$.

For any entity $e \in \mathcal{E}$ and relation $r \in \mathcal{R}$, this paper represents their corresponding text information as $l_e$ and $l_r$. For entity $e \in \mathcal{E}$, this paper regards the set $\mathcal{N}(e)$ of all its neighbor triplets as the context of $e$, specifically, $\mathcal{N}(e)$ is the union of the set $\{(e, r, e') \mid (e, r, e') \in \mathcal{G}, e' \in \mathcal{E}\}$ and the set $\{(e', r, e) \mid (e', r, e) \in \mathcal{G}, e' \in \mathcal{E}\}$, and for any neighbor triplet in $\mathcal{N}(e)$, We think that it expresses a piece of context information of node $e$.

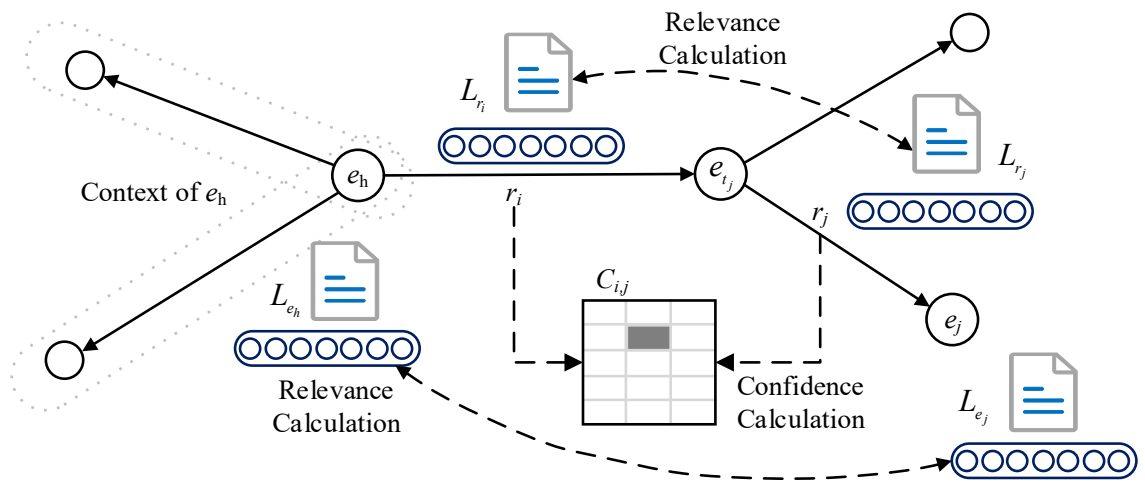

Fig. 1 An overview of the framework

Similar to Vashishth et al. [23], this paper also extends the relational set of knowledge map: $\mathcal{R} \leftarrow \mathcal{R} \cup \mathcal{R}_{\text{inverse}} \cup SL$, where $\mathcal{R}_{\text{inverse}} = \{r^{-1} \mid r \in \mathcal{R}\}$ is an inverse relational set. Specifically, for any triplet $(e_{\text{h}}, r, e_{\text{t}}) \in \mathcal{G}$, this paper adds an inverse relation $r^{-1}$ to the relational set. and correspondingly adds triplet $(e_{\text{t}}, r^{-1}, e_{\text{h}})$ to the knowledge graph $\mathcal{G}$, that is, $\mathcal{G} \leftarrow \mathcal{G} \cup \{(e_{\text{t}}, r^{-1}, e_{\text{h}})\}$. $SL$ stands for a set of self-ring relations, that is, for any entity $e \in \mathcal{E}$, a self-ring triple is added to the knowledge map $\mathcal{G}$, that is, $\mathcal{G} \leftarrow \mathcal{G} \cup \{(e, r_s, e)\}, r_{\text{s}} \in SL$. In addition, this paper uses $\mathcal{N}_e(e)$ represents the set of neighboring entities around entity $e$, and $\mathcal{N}_e(e)$ represents the set of neighboring relationships around Entity $e$. For example, for entity Beijing, its neighboring entity set is {North China, China, ...}, and the neighboring relationship set is located in, capital, introduction, English label, ...}.

3.2 Overall Model Architecture

The model proposed in this paper mainly consists of four parts: embedding layer, context encoding layer, feature fusion layer and decoding layer. First, the text is fed into the embedding layer to obtain the character-level embedding; then, the context features are extracted using Transformer and BiLSTM in the context encoding layer, and then fed into the feature fusion layer to be fused using the attention mechanism; finally, the conditional random fields are used in the decoding layer to decode and output the labels.

3.3 Embedding Layer

In this paper, we use Word2Vector and a pre-trained model as the embedding layer, in which the pre-trained model uses the RoBERTawwm model pre-trained by Xunfei Joint Laboratory of HITU.

Assuming that the initial input of the model is a sentence $\boldsymbol{S} = (x_1, x_2, \cdots, x_n)$. When using the RoBERTawwm model, the output is the character-level embedded human $\boldsymbol{R} = (r_1, r_2, \cdots, r_n)$ When using Word2Vector, the character-level embedding and binary character-level embedding are also obtained. When using Word2Vector word vectors, character-level embeddings and binary character-level embeddings are also obtained as $c = (c_1, c_2, \cdots, c_n)$ and $\boldsymbol{b} = (b_1, b_2, \cdots, b_n)$, where the character-level embedding is in word units, and the binary character-level embedding is in double-word units, which are stitched together to get the final word vector, as shown in equation (1).

$$\boldsymbol{V}_{\text{Vec}} = [\boldsymbol{c}; \boldsymbol{b}]$$

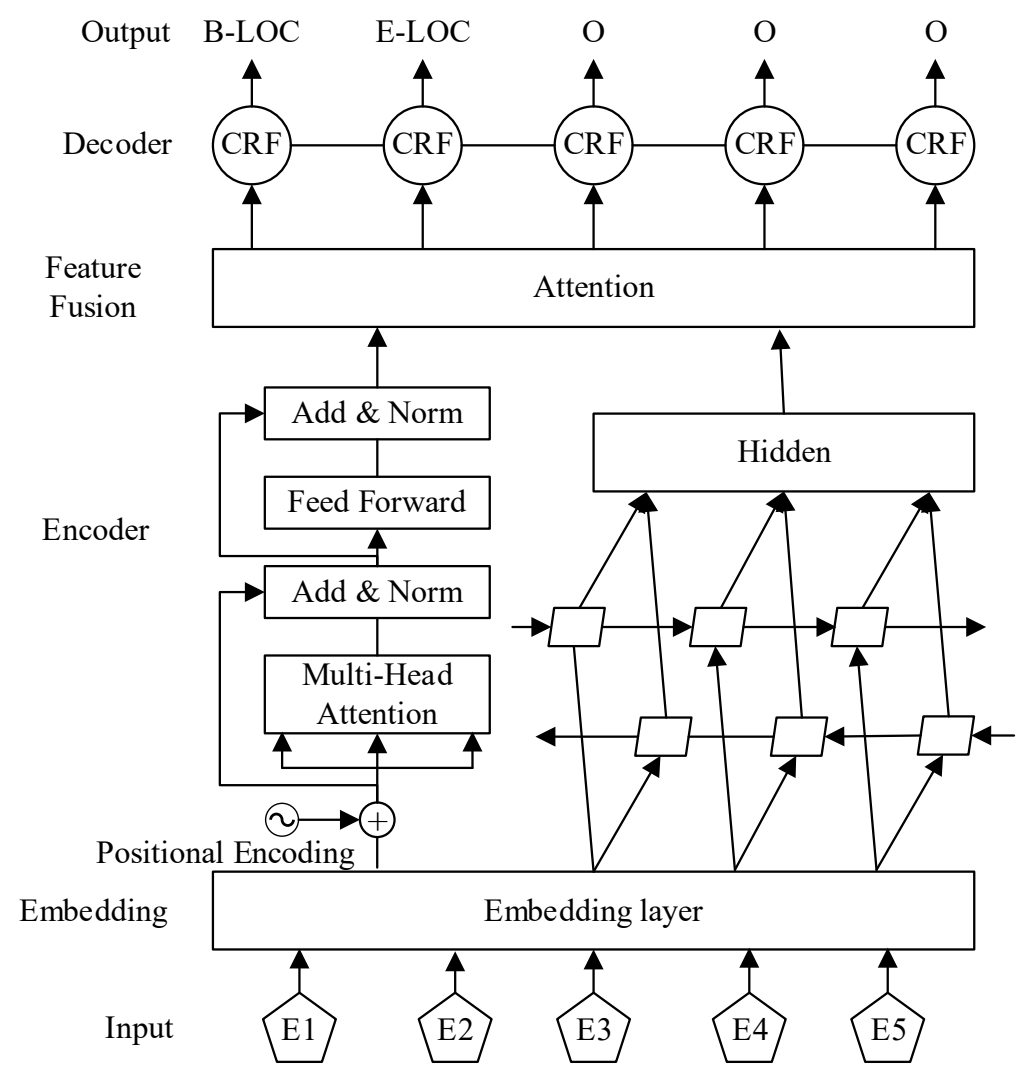

Fig. 2 The encoder-decoder structurte

3.4 Improved Transformer Encoder

In this paper, the left side of the context encoding layer is the structure of the Transformer encoder, which includes a multi-head self-attention layer, a feed-forward neural network layer, and uses layer normalization and residual concatenation. The original Transformer encoder employs absolute coding to generate the positional codes, The position code of the $t$ -th character is shown in (2):

$$P_{\text{PE},t,2i} = \sin\left(t/10000^{2i/d}\right)$$
$$P_{\text{PE},t,2i+1} = \cos\left(t/10000^{2i/d}\right)$$

where: The values range of $i$ is $\left[0, \frac{d}{2}\right]$; $d$ is the dimension of the input word vector. The resulting positional codes and word vectors are summed bitwise to obtain the input matrix of the multi-head self-attention layer $\boldsymbol{H} \in \mathbb{R}^{l \times d}$, where $l$ is the sequence length to be. $\boldsymbol{H}$ H is mapped to $\boldsymbol{Q}$, $\boldsymbol{K}$ and $\boldsymbol{V}$, as shown in (3):

$$\boldsymbol{Q}, \boldsymbol{K}, \boldsymbol{V} = \boldsymbol{H}\boldsymbol{W}_q, \boldsymbol{H}\boldsymbol{W}_k, \boldsymbol{H}\boldsymbol{W}_v$$

where: $\boldsymbol{W}_q$、 $\boldsymbol{W}_k, \boldsymbol{W}_v$ denotes the dimension of $\mathbb{R}^{d \times d_k}$ the variable weight matrix, the $d_k$ is the hyperparameter. The scaled dot product attention is computed by the following equation: the

$$A_{\text{Attention}}(\boldsymbol{K}, \boldsymbol{Q}, \boldsymbol{V}) = \text{Softmax}\left(\frac{\boldsymbol{Q}\boldsymbol{K}^{\text{T}}}{\sqrt{d_k}}\right)\boldsymbol{V}$$

When multiple self-attention is used, it is calculated as shown in Eq. (5) Eq. (7).

$$\boldsymbol{Q}^h, \boldsymbol{K}^h, \boldsymbol{V}^h = \boldsymbol{H}\boldsymbol{W}_q^h, \boldsymbol{H}\boldsymbol{W}_k^h, \boldsymbol{H}\boldsymbol{W}_v^h$$
$$\boldsymbol{D}^h = A_{\text{Attention}}(\boldsymbol{Q}^h, \boldsymbol{K}^h, \boldsymbol{V}^h)$$
$$M_{\text{Multi-Head}}(\boldsymbol{H}) = [\boldsymbol{D}^1, \boldsymbol{D}^2, \cdots, \boldsymbol{D}^h]\boldsymbol{W}_m$$

where: $h$ stands for Head index.; $[\boldsymbol{D}^1, \boldsymbol{D}^2, \cdots, \boldsymbol{D}^h]$ denotes the splicing of the attention of multiple Heads; the $\boldsymbol{W}_m$ denotes the dimension of $\mathbb{R}^{d \times d}$ of the variable weight matrix. The output of the multinomial self-attention layer $\boldsymbol{x}$ will be further processed by the feed-forward neural network layer, as shown in Eq. (8).

$$F_{\mathrm{FFN}}(\boldsymbol{x}) = \max(0, \boldsymbol{x}\boldsymbol{W}_1 + \boldsymbol{b}_1)\boldsymbol{W}_2 + \boldsymbol{b}_2$$

where: $\boldsymbol{W}_1$, $\boldsymbol{W}_2$, $\boldsymbol{b}_1$ and $\boldsymbol{b}_2$ are learnable parameters. $\boldsymbol{W}_1 \in \mathbb{R}^{d\times d_{ff}}, \boldsymbol{W}_2 \in \mathbb{R}^{d_{fj}\times d}, \boldsymbol{b}_1 \in \mathbb{R}^{d_{ff}}, \boldsymbol{b}_2 \in \mathbb{R}^{d}, d_{ff}$ is a hyper-parameter.

In this paper, the original Transformer encoder is improved by using relative position coding [16] and modifying the attention calculation . Firstly, the $\boldsymbol{H}$ maps to $Q, \mathbb{K}, , V, KK$ are not linearly transformed to break the symmetry and enhance the distance perception, and the transformation process is shown in Eq. (9).

$$\boldsymbol{Q}, \boldsymbol{K}, \boldsymbol{V} = \boldsymbol{H}\boldsymbol{W}_q, \boldsymbol{H}_{d_k}, \boldsymbol{H}\boldsymbol{W}_v$$

where. $\boldsymbol{W}_q, \boldsymbol{W}_q$ ldshendintweension of the $\mathbb{R}^{d\times d_k}, d_k$ is the feature dimension of one of the Heads; the $\boldsymbol{H}_{d_k}$ is the name of the $\boldsymbol{H}$ vector assigned to each Head.

Second, the absolute encoding of cos Functions are expressed in terms of sin function instead, the new relative position encoding is shown in equation (10).

$$\boldsymbol{R}_{t-j} = \left[\cdots \sin\left(\frac{t-j}{10000^{2i/d_k}}\right) \cos\left(\frac{t-j}{10000^{2i/d_k}}\right) \cdots\right]^{\mathrm{T}}$$

where : $t$ is the index of the target character $jj$ is the index of the context character ; $i$ The range of values is $\left[0, \frac{d_k}{2}\right]$. When calculating the attention score, the word vectors are calculated separately from the relative position encoding, and the bias term is added, and the calculation procedure is shown in Eq. (11).

$$\boldsymbol{A}_{\mathrm{rel},tj} = \boldsymbol{Q}_t\boldsymbol{K}_j^{\mathrm{T}} + \boldsymbol{Q}_t\boldsymbol{R}_{t-j}^{\mathrm{T}} + \boldsymbol{u}\boldsymbol{K}_j^{\mathrm{T}} + \boldsymbol{v}\boldsymbol{R}_{t-j}^{\mathrm{T}}$$

where : $\boldsymbol{Q}_t\boldsymbol{K}_j^{\mathrm{T}}$ denotes the attention fraction of the two characters; the $\boldsymbol{Q}_t\boldsymbol{R}_{t-j}^{\mathrm{T}}$ denotes the first $\boldsymbol{t}$ the deviation of individual characters in relative distance; the $\boldsymbol{u}\boldsymbol{K}_j^{\mathrm{T}}$ denotes the first $\boldsymbol{j}$ the deviation of the characters; the $\boldsymbol{v}\boldsymbol{R}_{t-j}^{\mathrm{T}}$ denotes the relative distance and direction bias term; the $\boldsymbol{u}$ and $v$ denotes the learnable parameters.

Finally, the attention is computed without scaling the dot product as shown in Equation (12).

$$A_{\text{Attention}}(\boldsymbol{Q}, \boldsymbol{K}, \boldsymbol{V}) = \operatorname{Softmax}(\boldsymbol{A}_{\text{rel}})\boldsymbol{V}$$

After the above modification of the attention , the position perception and orientation perception of the Transformer encoder are improved, which makes the Transformer suitable for the Chinese named entity recognition task.

3.5 Bidirectional Long Short-Term Memory Network

Long Short Term Memory Network (LSTM) is a special kind of Recurrent Neural Network (RNN), which can alleviate the problems of gradient vanishing and gradient explosion of traditional RNNs. In LSTM, a forgetting gate is introduced to control the information flow, so as to selectively memorize the information.

In the task, for the target character, this paper not only needs the information from above but also needs the information from below, therefore, BiLSTM is used as the context encoder, and its structure is shown on the right side of the context encoding layer of the overall model architecture in Fig. 2. BiLSTM adopts forward and backward inputs for the character-level embedding output from the embedding layer, and the forward and backward vectors are computed, and the two vectors are spliced together and used as the output of the hidden layer, which is realized as shown in Eq. (13) and Eq. (15).

$$\vec{\boldsymbol{h}}_t = \operatorname{LSTM}\left(\boldsymbol{x}_t, \overrightarrow{\boldsymbol{h}_{t-1}}\right)$$
$$\boldsymbol{h}_t = \operatorname{LSTM}\left(\boldsymbol{x}_t, \boldsymbol{h}_{t-1}\right)$$
$$\boldsymbol{h}_t = \left[\vec{\boldsymbol{h}}_t; \boldsymbol{h}_t\right]$$

3.6 Feature Fusion Layer

Transformer can model arbitrary distance dependencies, but it is not sensitive to position and orientation information; BiLSTM can fully capture orientation information, but cannot capture global information. In this paper, we borrow the gating mechanism and use the attention mechanism to dynamically fuse the context features extracted by the Transformer encoder and BiLSTM, so as to achieve the purpose of complementing each other's strengths. The dynamic fusion of attention mechanism is realized as shown in Eqs. (16) and (17).

$$\boldsymbol{z} = \sigma\left(\boldsymbol{W}_z^3 \tanh\left(\boldsymbol{W}_z^1\boldsymbol{x}_t + \boldsymbol{W}_z^2\boldsymbol{x}_b\right)\right)$$
$$\tilde{\boldsymbol{x}} = \boldsymbol{z} \cdot \boldsymbol{x}_t + (1 - \boldsymbol{z}) \cdot \boldsymbol{x}_b$$

where : $\boldsymbol{W}_z$ is a learnable weight matrix ; $\sigma$ is Sigmoid activation function. ; $\boldsymbol{x}_t$ is the vector of outputs from the Transformer encoder; the $\boldsymbol{x}_b$ is the vector of BiLSTM output. The vector $\boldsymbol{z}$ has the same dimension as $\boldsymbol{x}_t$ and $\boldsymbol{x}_b$ which is the same dimension as the weight between the two vectors, allows the model to dynamically determine how much information to use from the Transformer encoder or BiLSTM, thus remembering the important information and avoiding to cause an information light surplus.

3.7 Decoding Layer

In order to take advantage of the dependencies between different labels, this paper uses conditional random fields as the decoding layer. For a given sequence $\boldsymbol{s} = [s_1, s_2, \cdots, s_T]$, the corresponding label sequence is $\boldsymbol{y} = [y_1, y_2, \cdots, y_T]$ . $\boldsymbol{y}$ The probability is calculated as shown in equation (18).

$$P(\boldsymbol{y} \mid \boldsymbol{s}) = \frac{\sum_{t=1}^{T} \mathrm{e}^{f(y_{t-1}, y_t, s)}}{\sum_{y'}^{Y(s)} \sum_{t=1}^{T} \mathrm{e}^{f(y'_{t-1}, y'_t, s)}}$$

where : $f(\boldsymbol{y}_{t-1}, \boldsymbol{y}_t, \boldsymbol{s})$ denotes the computation of the distance from $\boldsymbol{y}_{t-1}$ to $\boldsymbol{y}_t$ The state transition fractions of $\boldsymbol{y}_t$ The fraction of the fraction, whose objective is $P(\boldsymbol{y} \mid \boldsymbol{s})$; $\boldsymbol{Y}(\boldsymbol{s})$ denotes all valid label sequences. When decoding, the Viterbi algorithm is used to find the globally optimal sequence.

## 4 Experiments

In this section, we firstly explain the dataset,comparison model and evaluation indexes used in theexperiments,and then present the experimental results of theproposedmodel and compare and analyze them with otherbenchmark models.

4.1 Introduction of datasets and comparison models

Table 1 Summary Statistics of Knowledge Graphs

| Dataset | FB15K-237 | WN18 |
|---|---|---|
| #Relation | 237 | 18 |
| #Entity | 14541 | 40943 |
| #Train | 271115 | 141442 |
| #Valid | 17535 | 2500 |
| #Test | 20466 | 2500 |

In this paper, experiments are conducted on two widely used data sets, namely FB15K-237 [29] and Wn18 [12], and their statistical data are shown in Table 1:

In order to verify the validity of the proposed model, this paper widely selects the knowledge map embedded learning model which has been widely used as the analogy method, including transition [11], distmult [30] and complex [24], R-gcn [22], KB Gan [31], Conve [26], ConvKB [32] SACN [23], Hyper [33], Rotate [34], ConVR [35], VR-GCN [36], CompGCN [21]. It has been introduced in detail above. Complex [24] is similar to Rescal [25] model and belongs to the model of link prediction based on matrix/tensor decomposition. R-GCN [22], VR-GCN [36] and COMPGCN [21] belong to the embedding representation model based on the graph convolution network. Taking R-GCN [22] as an example, it encodes the relationship in the knowledge graph into a matrix, and transmits the embedding information of adjacent entities through the relationship matrix, and adopts the multi-layer graph convolution network. KBGAN applies the Generative Adversary Network (Gan), which is generated in the training process. Conve [2] model is used as the decoder in this paper, and it is introduced in detail in Section 2. ConvKB [32], ConVR [35], SACN [23] and Hyper [33] are all methods based on convolutional neural networks. Take HypER [33] as an example, it can generate a simplified convolution filter related to relationships, and it can be constructed as tensor decomposition. Rotate [34] is similar to the translation mechanism-based model such as TRANSE [11], which represents the relationship between entities as the rotation from entity to entity in vector space.

4.2 Evaluation Methodology Description

In this paper, the validity of the model is evaluated by linking the prediction tasks. In the experiment, for the test triplets whose heads or tails have been turned off in advance, this paper speculates the head or tail entities that have been removed based on the learned embedding representation. For each test triplet, this paper selects any entity in the knowledge map as the possible prediction result, and calculates the score value after completing the test triplet with this entity, as shown in Equation (12). Finally, the score values are sorted. Here, taking the prediction of missing header entities as an example, for each triplet $(e_{\mathrm{h}}, r, e_{\mathrm{t}})$ in the test set, the header entity $e_{\mathrm{h}}$ is deleted in advance, and then any entity $e_{\mathrm{hc}} \in \mathcal{E}$ in $G$ is tried to complete the test triplet, thus generating a set of candidate triplets $\{(e_{\mathrm{hc}}, r, e_{\mathrm{t}}) \mid e_{\mathrm{hc}} \in \mathcal{E}\}$. Based on the learned embedded representation, the scores of candidate triples are calculated and sorted. The higher the scores, the more reliable the learned model, that is, the embedded representation. By comparing with the real results, the quality of the learned embedded representation can be judged.

Finally, MR(mean rank), MRR (mean reciprocal rank) and Hit@k are used as evaluation indicators [12]. Among them, both MR and MRR are indicators of the average ranking of prediction results, and Hit@k refers to the proportion of prediction results in the top K, and this paper specifically adopts Hit@10, Hit@3 and Hit@1. In short, the better.

4.3 Experimental Setup

The experimental code in this paper is implemented in Python, and it is completed on the server with Ubuntu16.04.6 LTS operating system. Its CPU configuration is 16-core Intel Core i7-6900K 3.20 GHz, and its memory is 128 GB. The GPU configuration is 4 GeForce GTX 1080 GPU cards.

For the encoding of text representation vectors of entities and relationships, this paper uses the pre-trained-Bert-base-uncased model 0, the initial dimension of text vectors is 768, and the transformed dimension is 200. In the graph-convolution network, the initialization vector dimension of entities and relationships is 100, that is, d=100, and the dimension of GCN is 200. That is, $d' = 200$. The height and width of dimension transformation in the decoder are 10 and 20 respectively, and the size of convolution filter is 7×7, and the number is 200. Adam optimizer is used to train the whole model, and the batch size is 256 and the learning rate is 0.001.

In this paper, the TransE model is reproduced, and the other models refer to the results reported in the comparative model paper.

4.4 Analysis of Experimental Results

Table 2 reports the experimental results of this model and the comparison model for the link prediction task.

The following results can be observed from Table2:

1) Our model significantly outperforms benchmark methods such as TransE, DistMult, and ComplEx across all evaluation metrics, and achieves performance comparable to recently proposed models such as SACN, HypER, and CompGCN, thereby demonstrating its effectiveness. On the FB15K-237 dataset, our model ranks first in the Hit@10 index.

2) For Hit@1 and Hit@3, the performance gap between our model and strong baselines such as CompGCN, ConvR, and SACN is minimal. Specifically, the Hit@1 score is only 1.51% lower than the best-performing CompGCN, while the MRR score is just 0.8% lower. On the WN18 dataset, our model ranks first in the MR index, and the differences in Hit@10 and Hit@3 compared to the top models are also very small—only 0.2% lower than RotatE in Hit@10, and just 0.9% lower than ConvR and HypER in Hit@3.

3) In general, embedding learning methods based on graph neural networks outperform traditional models such as TransE that rely solely on structural information. In our case, the proposed model achieves substantial performance gains in link prediction tasks by jointly embedding contextual and textual information into the knowledge graph through graph convolutional networks.

Table2 Link Prediction Results on FB15K-237 and WN18

| Model | FB15K-237 | | | | | WN18 | | | | |
|---|---|---|---|---|---|---|---|---|---|---|
| | MRR | MR | Hit@10 | Hit@3 | Hit@1 | MRR | MR | Hit@10 | Hit@3 | Hit@1 |
| TransE[12] | 0.294 | 357 | 0.330 | 0.330 | 0.146 | 0.454 | 251 | 0.891 | 0.803 | 0.064 |
| DistMul[30] | 0.241 | 254 | 0.419 | 0.263 | 0.155 | 0.829 | | 0.829 | 0.923 | 0.726 |
| ComplEx[24] | 0.247 | 339 | 0.428 | 0.275 | 0.158 | | 844 | 0.940 | | |
| R-GCN[22] | 0.248 | | 0.417 | | 0.151 | 0.773 | | 0.944 | 0.889 | 0.65 |
| KBGAN[31] | 0.278 | | 0.458 | | | 0.779 | | 0.949 | | |
| ConvE [26] | 0.325 | 244 | 0.501 | 0.356 | 0.237 | 0.943 | 374 | 0.956 | 0.946 | 0.935 |
| ConvKB[30] | 0.243 | 311 | 0.421 | 0.371 | 0.155 | | | | | |
| SACN[23] | 0.350 | | **0.540** | **0.390** | **0.260** | | | | | |
| HypER[33] | 0.341 | 250 | 0.520 | 0.376 | 0.252 | 0.951 | 431 | 0.958 | 0.955 | 0.947 |
| RotatE[34] | 0.338 | **177** | 0.533 | 0.375 | 0.241 | | 309 | **0.959** | | |
| ConvR[35] | 0.350 | | 0.528 | 0.385 | 0.261 | **0.951** | | 0.958 | **0.955** | **0.947** |
| VR-GCN[36] | 0.248 | | 0.432 | 0.272 | 0.159 | 0.847 | | 0.946 | 0.929 | 0.764 |
| CompGCN[21] | **0.355** | 197 | 0.535 | 0.390 | 0.264 | | | | | |
| Our Methods | 0.352 | 186 | 0.536 | 0.385 | 0.385 0.260 | 0.928 | **240** | 0.957 | 0.946 | 0.909 |

Note: The best performance is in bold.

## 5 Conclusion

This paper addresses the limitations of current knowledge map embedding learning methods, which typically focus on the structure information in triplets while overlooking the contextual and textual data within the knowledge map. To enhance link prediction and similar tasks, we introduce a novel approach utilizing a graph convolutional neural network, integrating context and textual information into the learning of embedding representations. To emphasize the significance of context at a finer granularity, we establish an efficient rule for gauging context confidence. Additionally, we devise a technique for assessing context relevance through vector representations of textual data, thereby augmenting the influence and guidance of context information. The efficacy of our model is demonstrated through comparative experiments on two extensively utilized benchmark datasets.